\pdfoutput=1

\documentclass[11pt]{article}

\usepackage[final]{acl}

\usepackage{times}
\usepackage{latexsym}

\usepackage[T1]{fontenc}

\usepackage[utf8]{inputenc}

\usepackage{microtype}
\usepackage{CJKutf8}

\usepackage{inconsolata}
\usepackage{tabularx}
\usepackage{arydshln}
\usepackage{graphicx}
\usepackage{enumitem}
\usepackage{bm}
\usepackage{float}
\restylefloat{table}
\restylefloat{figure}
\usepackage{adjustbox}
\usepackage{multirow}
\usepackage{booktabs}
\usepackage{enumitem}
\usepackage{algorithm}
\usepackage{algpseudocode}
\usepackage{amsmath}
\usepackage[normalem]{ulem}
\usepackage{array}
\usepackage{booktabs}
\usepackage{tikz}
\usepackage{amssymb}
\usepackage{amsfonts}

\title{LLMFactor: Extracting Profitable Factors through Prompts for Explainable Stock Movement Prediction}

\author{Meiyun Wang \thanks{Corresponding Author} \\ The University of Tokyo \\ omiun20@g.ecc.u-tokyo.ac.jp 
        \And
        Kiyoshi Izumi \\ The University of Tokyo \\ izumi@sys.t.u-tokyo.ac.jp
        \And
        Hiroki Sakaji \\ Hokkaido University \\ sakaji@ist.hokudai.ac.jp}

\begin{document}
\maketitle
\begin{abstract}

Recently, Large Language Models (LLMs) have attracted significant attention for their exceptional performance across a broad range of tasks, particularly in text analysis. However, the finance sector presents a distinct challenge due to its dependence on time-series data for complex forecasting tasks. In this study, we introduce a novel framework called LLMFactor, which employs Sequential Knowledge-Guided Prompting (SKGP) to identify factors that influence stock movements using LLMs. Unlike previous methods that relied on keyphrases or sentiment analysis, this approach focuses on extracting factors more directly related to stock market dynamics, providing clear explanations for complex temporal changes. Our framework directs the LLMs to create background knowledge through a fill-in-the-blank strategy and then discerns potential factors affecting stock prices from related news. Guided by background knowledge and identified factors, we leverage historical stock prices in textual format to predict stock movement. An extensive evaluation of the LLMFactor framework across four benchmark datasets from both the U.S. and Chinese stock markets demonstrates its superiority over existing state-of-the-art methods and its effectiveness in financial time-series forecasting.
\end{abstract}

\section{Introduction} 
Artificial Intelligence (AI) has become a staple in the financial sector, addressing various challenges such as predicting stock movements \cite{patel2015predicting, henrique2019literature}, providing robo-advisory services \cite{xue2018group, bertrand2023questioning}, and managing risks \cite{ahbali-etal-2022-identifying, wang2021deeptrader}. Among these tasks, forecasting stock trends is particularly significant, as it leverages historical data to shape trading strategies and pinpoint opportunities for buying or selling stocks.

The Efficient Market Hypothesis (EMH), proposed by Eugene Fama \cite{fama1970efficient}, suggests that stock prices reflect all available information, making it difficult to predict future price movements. However, subsequent research has identified limitations to market efficiency, highlighting how phenomena such as information asymmetry \cite{timmermann2004efficient} and irrational behaviours \cite{lo2004adaptive} can induce departures from perfect efficiency. These observations have paved the way for researchers to seek excess market returns by identifying and leveraging market inefficiencies. 

\begin{figure}[t]
  \centering
  \includegraphics[width=\linewidth]{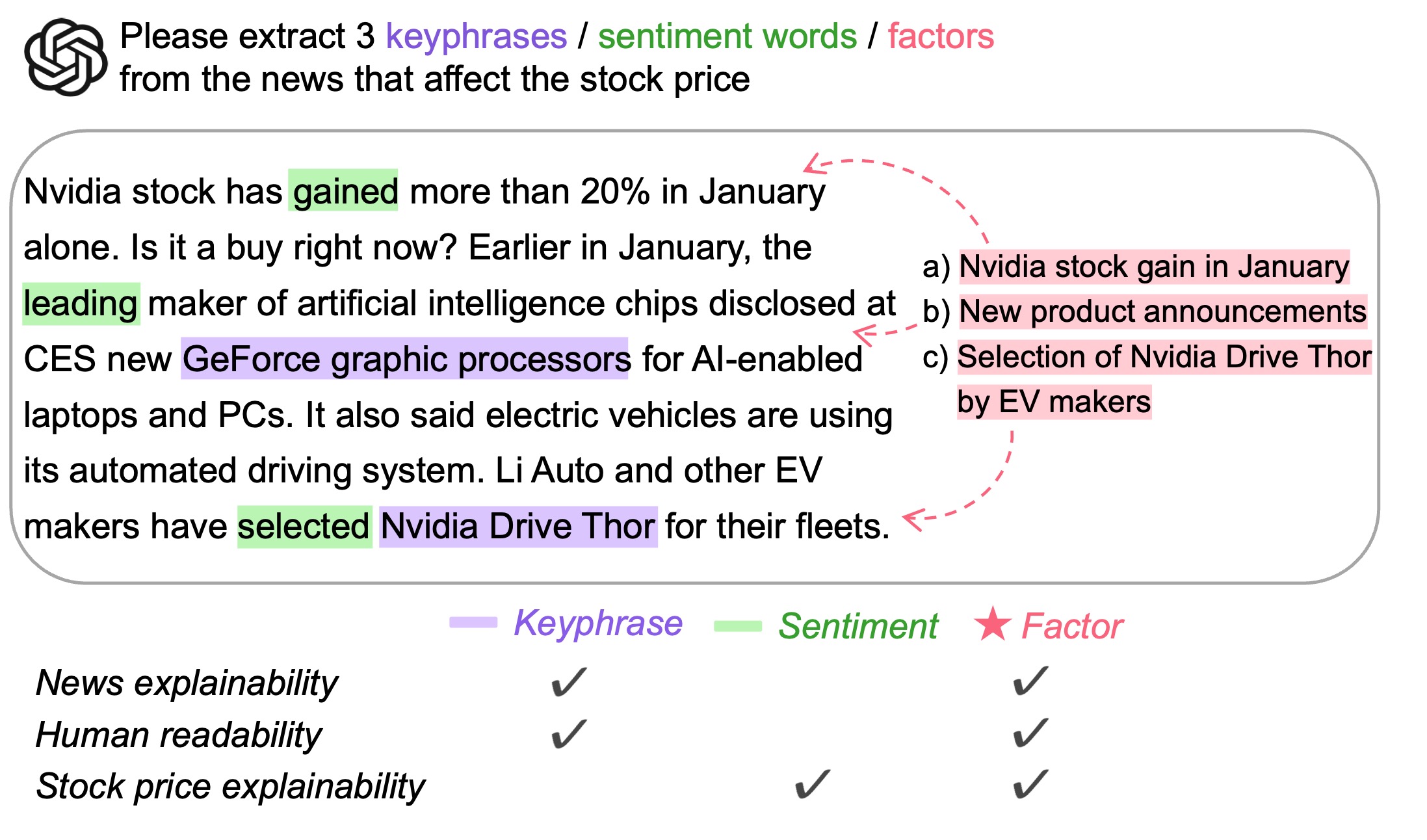}
  \caption{Comparing keyphrases, sentiment words, and factors: LLM-extracted factors offer greater human readability and explainability.}
  \label{fig: outline}
\end{figure}
Building on this foundation, there is a growing interest in exploring various data types to enhance prediction capabilities. Several studies underscore the importance of news related to stocks in uncovering fundamental market insights \cite{vargas2018deep, li2021modeling}, while others highlight the significance of understanding the interconnections between companies and industries \cite{feng2019temporal, hsu2021fingat}. Recent research has empirically demonstrated the impact of public sentiment on market trends, with efforts to extract sentiment and keyphrases from news and social media data \cite{nguyen2015sentiment, hao2021predicting}. 

However, as illustrated in Figure \ref{fig: outline}, these methods encounter various limitations. We propose a novel task to enhance stock movement prediction through the use of \textbf{``factors.''} Figure \ref{fig: outline} delineates the distinctions among keyphrases, sentiment words, and factors generated by LLMs. For the given stock-related news, keyphrases may outline the content but do not directly correlate with the stock price. Conversely, sentiments are associated with stock price but lack clarity for human interpretation. Consequently, factors offer superiority in three aspects: \textbf{news explainability, human readability, and stock price explainability.} 

To incorporate factors into financial forecasting, we introduce a new framework \textbf{LLMFactor}, which derives factors from LLMs through \textbf{Sequential Knowledge-Guided Prompting (SKGP)} and then explains stock price trends. LLMFactor represents a holistic approach to integrating LLMs into financial applications. Initially, our proposed SKGP strategy prompts LLMs to generate factors related to stocks. SKGP starts with a fill-in-the-blank technique to elicit background knowledge (e.g. company relations) related to the stock. Subsequently, we prompt LLMs to identify reliable factors from news articles. Finally, we compile the text-formatted time-series data with these factors and knowledge to predict stock price trends. Experiments across four datasets indicate that LLMFactor yields superior predictive results and can clarify the principles behind these predictions. Additionally, the obtained factors significantly help to dynamically present market changes over time in a format that is accessible and understandable to humans.

The main contributions of this study are summarized as follows:
\begin{itemize}[noitemsep,leftmargin=*,align=left]
    \item \textbf{New Task - Factor Extraction:} We introduce a novel task, factor extraction, aimed at extracting significant factors from textual data to assist in predicting time-series data. This task surpasses traditional keyphrase-based and sentiment-based methods in three key areas: the explainability of news content, the readability for humans, and the ability to explain stock price movements.
    \item \textbf{New Strategy - SKGP:} We propose Sequential Knowledge-Guided Prompting (SKGP) as an innovative strategy for leveraging background knowledge in the process of stock movement prediction. Unlike basic prompting methods, SKGP employs a fill-in-the-blank approach, utilizing minimal background knowledge to enhance the richness of the prompt template.
    \item \textbf{New Framework - LLMFactor:} 
    The LLMFactor framework, developed in this study, utilizes factors derived from LLMs to elucidate the dynamic temporal changes. This framework has proven to be a valuable tool in financial applications, providing deep insights into market trends and facilitating factor analysis.
\end{itemize}

\section{Related Work}
\subsection{Stock Movement Prediction using Textual Data}
With the advancement of natural language processing (NLP) techniques, many researchers leverage textual data to forecast stock market trends. \cite{xu-cohen-2018-stock} uses tweets and historical prices to make temporally dependent predictions from stock data. \cite{luo-etal-2023-causality} models the multi-modality between financial text data and causality-enhanced stock correlations. These efforts aim to develop sophisticated models to enhance the prediction accuracy and consider the textual data as a whole.

Furthermore, certain studies focus on extracting more granular insights from textual data. \cite{zhou-etal-2021-trade} identifies corporate events as the driving force behind stock movements while \cite{wan2021sentiment} demonstrates the significant association between strong media sentiment and market return. \cite{wang2020incorporating} enhances stock movement prediction by including expert opinions aggregated from various sources. These insights offer greater clarity to the understanding of market dynamics.

\begin{figure*}
  \centering
  \includegraphics[width=\linewidth]{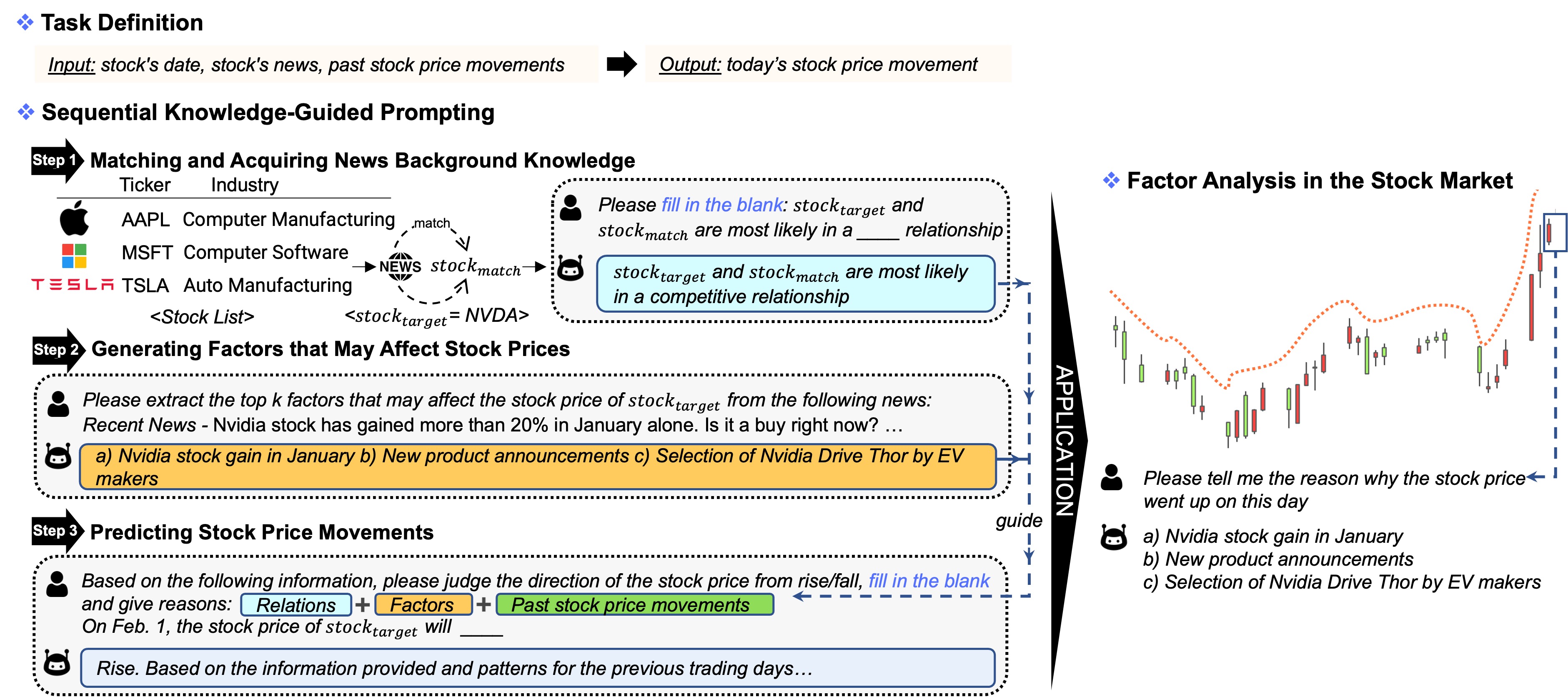}
  \caption{LLMFactor: From Sequential Knowledge-Guided Prompting (SKGP) to factor analysis in the stock market. The SKGP technique comprises three steps that sequentially generate knowledge and predict stock movement, guided by this knowledge. In applying factor analysis to stock movements, factors offer timely and concise explanations for their changes.}
  \label{fig: overall}
\end{figure*}

\subsection{Time-series Forecasting with LLMs}
LLMs, such as variants of the Generative Pre-trained Transformer (GPT), are constructed from wide-ranging and diverse datasets, significantly enriching their extensive knowledge base \cite{brown2020language}. Yet, their foundational structure, grounded in the transformer architecture, is not intrinsically designed for analyzing time-series data \cite{gruver2023large}. This limitation has sparked increased interest among scholars in exploring methods to tailor LLMs for proficient time-series prediction.

\cite{xue2023promptcast} proposes a prompt-based approach to transform numerical inputs and outputs into textual prompts, framing the forecasting task as a sentence-to-sentence conversion. This method allows language models to be directly applied to forecasting tasks. For financial time series forecasting, \cite{yu-etal-2023-harnessing} employs prompts to generate summaries and keyphrases from LLMs, incorporating multiple data sources to enhance time-series forecasting. Their approaches offer valuable insights into stock market trends. However, their prompts contain excessive information, resulting in responses from LLMs that are relatively lacking in detail.

\subsection{Prompt Engineering}
Prompt engineering allows LLMs to address a variety of tasks effectively through carefully crafted prompts. However, identifying the optimal prompt for each specific task is a challenge \cite{liu2023pre}. This challenge leads to the exploration of diverse prompting strategies. For instance, \cite{wei2022chain} introduces a chain of thought (CoT) approach, which involves a sequence of intermediate reasoning steps to enhance the capability of LLMs in complex reasoning tasks. Meanwhile, \cite{liu-etal-2022-generated} proposes a generated knowledge prompting technique, which involves generating knowledge through a language model and then using this generated knowledge as supplementary input for answering questions. Another popular technique, known as Retrieval-Augmented Generation (RAG) \cite{lewis2020retrieval}, incorporates external knowledge into LLMs through retrieve-augment-generate flows. Inspired by these techniques, we propose a novel Sequential Knowledge-Guided Prompting, designed to enhance the accuracy of financial forecasts.

\section{LLMFactor}
The proposed LLMFactor is a comprehensive framework to predict and explain stock market trends as shown in Figure \ref{fig: overall}. 

\subsection{Task Definition}
For a given stock, denoted as $stock_{target}$, we consider its associated news published on the target prediction date $date_{target}$, referred to as $news_{target}$, and its historical stock price sequence $P=\{P_1, P_2,..., P_t\}$, where $t$ represents the window size. The task of predicting stock movement is formulated as a binary classification problem, where the stock price sequence is transformed into a series of stock movements, $\hat{P}=\{\hat{P_1}, \hat{P_2},..., \hat{P_t}\}$. In this series, $\hat{P_i}=1$ indicates a rise in the stock price from $P_{i-1}$ to $P_i$, whereas $\hat{P_i}=0$ indicates a fall. Our objective is to predict $\hat{P}_{t+1}$ given $date_{target}$, $news_{target}$, and $\hat{P}$.

\subsection{Sequential Knowledge-Guided Prompting}
\subsubsection{Matching and Acquiring News Background Knowledge}
The foundation of our approach is the Sequential Knowledge-Guided Prompting (SKGP) strategy, which comprises three primary stages. Consider Nvidia (NVDA) as our example $stock_{target}$, the initial phase involves matching the stock with relevant news and acquiring background knowledge. Let $S = \{(C_i, T_i, I_i) \mid i \in \mathbb{N}, i \leq n \}$ be the stock list, where each tuple $(C_i, T_i, I_i)$ consists of a company $C_i$, its ticker symbol $T_i$, and its industry $I_i$. These tuples are collected from the NYSE and NASDAQ exchanges, with $n$ being a positive integer representing the total number of such tuples. We match $S$ with $news_{target}$ to obtain $stock_{match} = \{ S \cap news_{target}\}$. Following this, we prompt LLMs to obtain relations between $stock_{target}$ and $stock_{match}$. The prompting method is defined as 
$LLM: String \rightarrow String$, where the input is a $RelationTemplate$ ``Please fill in the blank: $stock_{target}$ and $stock_{match}$ are most likely in a \_\_\_ relationship,'' and the output is the type of relationship, denoted as $relation$.

This method of deriving background knowledge about $news_{target}$ significantly improves our understanding of the news content. For example, if $stock_{target}$ and $stock_{match}$ are identified as competitors, information about $stock_{match}$ could potentially have a negative impact on $stock_{target}$. Furthermore, numerous studies have underscored the critical role of company relationships in predicting stock market movements \cite{hsu2021fingat, ang-lim-2022-guided}. Therefore, our fill-in-the-blank technique aims to constrain the response format, facilitating a direct and unambiguous identification of the relationship.

\subsubsection{Generating Factors that May Affect Stock Prices}
The next step for SKGP involves generating factors from \(news_{target}\). The importance of these factors is threefold: 1) They are more closely associated with stock movements than keyphrases, sentiments, news summaries, or entire news articles, thus providing a higher likelihood of profitable market trend predictions. 2) Factors derived from news texts offer more immediate and detailed insights into stock price fluctuations compared to those obtained from other sources. 3) They improve the explainability of stock price trends and the rationale behind LLMs' predictions.

To generate reliable factors, we instruct LLMs to analyze the news content and identify factors that could influence the stock price. This approach fully utilizes the intrinsic knowledge of LLMs. Considering the top $k$ factors we aim to extract for the given \(stock_{target}\) and \(news_{target}\), the prompting method is described as \(LLM(FactorTemplate) = factor\), where \(FactorTemplate\) is a structured sentence: ``Please extract the top $k$ factors that may affect the stock price of \(stock_{target}\) from the following news,'' followed by \(news_{target}\), and the output is the factors generated by the LLMs. The factors produced by LLMs are not restricted to the words found in the news; instead, LLMs consider the content of the news and its potential impact on stock movement, often summarizing the significant elements in the content. For example, the factor ``Nvidia stock gain in January'' mentioned in Figure \ref{fig: overall} suggests that the event was Nvidia's past stock performance, and the impact was a gain, which could positively influence Nvidia's future stock price.

\subsubsection{Predicting Stock Price Movements}
To predict stock movement, we integrate news background knowledge and factors to guide LLMs. Meanwhile, we transform time-series data into a textual format for LLMs to understand. In a sequence of stock movements denoted by $\hat{P}=\{\hat{P_1}, \hat{P_2},...,\hat{P_t}\}$, we introduce a function where $f(\hat{P_i}=0)$ is assigned the value ``fell'' and $f(\hat{P_i}=1)$ is assigned ``rose,'' thereby converting $\hat{P}$ into a sequence of outcomes $TextMovement = \{ f(\hat{P_i}) \mid i \in \{1, 2, \ldots, t\}, f(\hat{P_i}) = \text{``fell''} \text{ if } \hat{P_i} = 0 \text{ and } \text{``rose''} \text{ if } \hat{P_i} = 1 \}$. Given the textual stock movement sequence $TextMovement$ and its date series $date=\{date_1, date_2,...,date_t\}$, the past stock price movements are transformed into a $TimeTemplate$, structured as ``On $date_i$, the stock price of $stock_{target}$ $f(\hat{P_i})$.''

Subsequently, we construct a $PriceTemplate$ that includes an initial instruction, ``Based on the following information, please judge the direction of the stock price as rise or fall, fill in the blank and give reasons,'' followed by a concluding instruction, ``On $date_i$, the stock price of $stock_{target}$ will \_\_\_.'' By integrating the $relation$, $factor$, $TimeTemplate$, and $PriceTemplate$, we articulate the prompting method as $LLM(relation, factor, TimeTemplate,$ $PriceTemplate) = prediction.$ The $prediction$ outcome specifies whether the stock price will ``rise'' or ``fall,'' along with the rationale for this inference. More details of templates are in Appendix \ref{appendix:templates of skgp}.
\begin{table*}[h]
\centering
\caption{Statistics of four benchmark datasets: StockNet, CMIN-US, and CMIN-CN show consistent processing across each dataset, with every stock represented by time-series data that includes both tweets and stock prices. In contrast, the EDT dataset is composed of news articles, with each article linked to the relevant stock and its corresponding stock price.}\label{tbl:dataset}
\begin{adjustbox}{width=\textwidth}
\begin{tabular}{ccccc}
\hline
\textbf{Statistics} & \textbf{StockNet} & \textbf{CMIN-US} & \textbf{CMIN-CN} & \textbf{EDT}  \\ \hline \hline
\multicolumn{1}{c|}{Data Type} & \multicolumn{1}{c|}{time series \& text} & \multicolumn{1}{c|}{time series \& text} & \multicolumn{1}{c|}{time series \& text} & \multicolumn{1}{c}{text}  \\ \hline
\multicolumn{1}{c|}{Data Resource} & \multicolumn{1}{c|}{price sequence \& tweets} & \multicolumn{1}{c|}{price sequence \& tweets} & \multicolumn{1}{c|}{price sequence \& tweets} & \multicolumn{1}{c}{prices \& news articles}  \\ \hline
\multicolumn{1}{c|}{Data Size} & \multicolumn{1}{c|}{19,318} & \multicolumn{1}{c|}{83,553} & \multicolumn{1}{c|}{198,781} & \multicolumn{1}{c}{54,080}  \\ \hline
\multicolumn{1}{c|}{Stock Market} & \multicolumn{1}{c|}{US} & \multicolumn{1}{c|}{US} & \multicolumn{1}{c|}{CN} & \multicolumn{1}{c}{US}  \\ \hline
\multicolumn{1}{c|}{Stock Number} & \multicolumn{1}{c|}{87} & \multicolumn{1}{c|}{110} & \multicolumn{1}{c|}{300} & \multicolumn{1}{c}{4,228}  \\ \hline
\multicolumn{1}{c|}{Date Range} & \multicolumn{1}{c|}{2014-01-01 to 2016-01-01} & \multicolumn{1}{c|}{2018-01-01 to 2021-12-31} & \multicolumn{1}{c|}{2018-01-01 to 2021-12-31} & \multicolumn{1}{c}{2020-03-01 to 2021-05-06}  \\ \hline
\end{tabular}
\end{adjustbox}
\end{table*}

\subsection{Factor Analysis in the Stock Market}
The SKGP presents a powerful technique for predicting stock movements, and the factors derived from SKGP offer additional insights into stock market trends. For instance, as illustrated in Figure \ref{fig: overall}, factor analysis can be applied to the stock market. Taking Nvidia's stock price trend as an example, after a consistent increase over the past five days, the day highlighted with a blue box also exhibits an upward movement. To explain this occurrence, our LLMFactor identifies a concise set of factors, such as ``Nvidia stock gain in January, new product announcements, and selection of Nvidia Drive Thor by EV makers.''

\section{Experiments}
By conducting experiments on benchmark datasets, we aim to address the following research question: Does LLMFactor offer improved prediction accuracy and more insightful explanations for stock market movements?
\subsection{Datasets}
We evaluate LLMFactor on four benchmark datasets, in alignment with the current SOTAs: a) \textbf{StockNet} \cite{xu-cohen-2018-stock} includes 87 stocks from 9 industries, accompanied by stock-related tweets and historical price data from 2014-01-01 to 2016-01-01, in the US stock market; b) \textbf{CMIN-US} \cite{luo-etal-2023-causality} comprises the top 110 stocks along with their tweets and historical price data from 2018-01-01 to 2021-12-31, in the US stock market; c) \textbf{CMIN-CN} \cite{luo-etal-2023-causality} consists of the 300 stocks in the CSI300 index \footnote{A market capitalization-weighted index crafted to mirror the performance of the leading 300 stocks traded on the Shanghai Stock Exchange and the Shenzhen Stock Exchange.}, with their tweets and historical price data from 2018-01-01 to 2021-12-31, in the Chinese stock market; d) \textbf{EDT} \cite{zhou-etal-2021-trade} includes 54,080 news articles from 2020-03-01 to 2021-05-06, along with related stock and stock price information in the US stock market. Unlike StockNet, CMIN-US, and CMIN-CN, which focus on time-series forecasting, the EDT dataset is centred on news content. The statistics of all datasets are summarized in Table \ref{tbl:dataset}.

\subsection{Evaluation Metrics}
Following previous studies by \cite{xu-cohen-2018-stock, luo-etal-2023-causality, zhou-etal-2021-trade}, we apply Accuracy (ACC) and Matthews Correlation Coefficient (MCC) as the evaluation metrics. Given the confusion matrix $\left[\begin{smallmatrix} \text{tp} & \text{fp} \\ \text{fn} & \text{tn} \end{smallmatrix}\right]$, ACC and MCC are defined as:\\
\noindent
\begin{minipage}{\linewidth} 
\small 
\begin{equation}
    \text{ACC} = \frac{tp + tn}{tp + tn + fp + fn}
\end{equation}
\begin{equation}
    \text{MCC} = \frac{tp \times tn - fp \times fn}{\sqrt{(tp+fp)(tp+fn)(tn+fp)(tn+fn)}}
\end{equation}
\end{minipage}
\\
\subsection{Baselines}
\noindent \textit{Keyphrase-based.} We employ several models to identify keyphrases in text for prediction purposes:
\begin{itemize}[noitemsep,leftmargin=*,align=left]
    \item \textbf{PromptRank}\cite{kong-etal-2023-promptrank}: An unsupervised approach that utilizes pre-trained language models (PLMs) with a focus on leveraging prompt-based techniques for keyphrase extraction.
    \item \textbf{KeyBERT}\cite{grootendorst2020keybert}: Leverages BERT embeddings to identify key sub-phrases in documents through cosine similarity, emphasizing the most representative phrases.
    \item \textbf{YAKE}\cite{campos2020yake}: A lightweight unsupervised method for keyword extraction, employing statistical features from individual documents to pinpoint the most pertinent keywords.
    \item \textbf{TextRank}\cite{mihalcea-tarau-2004-textrank}: An unsupervised approach identifies keyphrases based on the co-occurrence graph of words in a text.
    \item \textbf{TopicRank}\cite{bougouin-etal-2013-topicrank}: Groups similar expressions into topics and ranks these topics to identify keyphrases.
    \item \textbf{SingleRank}\cite{wan-xiao-2008-collabrank}: An extension of TextRank that weights edges in the co-occurrence graph based on the number of co-occurrences.
    \item \textbf{TFIDF}\cite{sparck1972statistical}: A traditional unsupervised approach that evaluates the importance of a word within a document in a corpus.
\end{itemize}

\noindent \textit{Sentiment-based.} We employ various models to analyze sentiment in texts and predict stock market movements:
\begin{itemize}[noitemsep,leftmargin=*,align=left]
    \item \textbf{EDT}\cite{zhou-etal-2021-trade}: Utilizes a bi-level event detection framework based on BERT to classify corporate events impacting stock prices.
    \item \textbf{FinGPT}\cite{zhang2023instructfingpt}: Employs instruction tuning on LLMs to enhance financial sentiment analysis capabilities.
    \item \textbf{GPT-4-turbo}: Builds upon GPT-4's capabilities, enhancing efficiency and performance, especially for processing longer texts.
    \item \textbf{GPT-4}: Represents a significant leap forward from previous iterations, with improved processing speeds and enhanced handling of extended text inputs.
    \item \textbf{GPT-3.5-turbo}: Known for its comprehensive understanding and human-like text generation, with optimized processing for multiple tasks.
    \item \textbf{RoBERTa}: Applies Financial RoBERTa\footnote{\url{https://huggingface.co/soleimanian/financial-roberta-large-sentiment}} for sentiment analysis in English texts and a fine-tuned sentiment-specific RoBERTa model\cite{Wang2022Fengshenbang1B} for Chinese texts.
    \item \textbf{FinBERT}: A BERT model fine-tuned for financial sentiment classification, available in both English \cite{araci2019finbert} and Chinese versions\footnote{\url{https://huggingface.co/bardsai/finance-sentiment-zh-base}}.
\end{itemize}
\noindent \textit{Time-based.} We employ models that leverage both textual and time-series data to predict stock movement:
\begin{itemize}[noitemsep,leftmargin=*,align=left]
    \item \textbf{CMIN}\cite{luo-etal-2023-causality}: An end-to-end deep neural network that models the multimodality between financial text data and causality-enhanced stock correlations.
    \item \textbf{StockNet}\cite{xu-cohen-2018-stock}: A deep generative model that jointly exploits textual and price signals for stock prediction.
\end{itemize}

\subsection{Implementation Details}
For keyphrase-based methods, the datasets are divided into two subsets based on labels: one containing data labeled with \(0\) (indicating a fall) and the other with data labeled with \(1\) (indicating a rise). Keyphrase extraction is applied to each subset to identify negative and positive keyphrases for each stock. The score for a given text is calculated as follows:
Let $POS$ be the set of positive keyphrases and $NEG$ be the set of negative keyphrases. For a piece of text $T$, the score $S$ is computed as:
\noindent
\begin{minipage}{\linewidth}
\small 
\begin{equation}
    S = \sigma\left( \sum_{p \in POS} \mathbb{I}(p \in T) - \sum_{n \in NEG} \mathbb{I}(n \in T) \right) 
\end{equation}
\end{minipage}
\\
where \( \sigma(x) \) is the sigmoid function defined as \( \sigma(x) = \frac{1}{1 + e^{-x}} \), and \( \mathbb{I}(\cdot) \) is an indicator function that equals \(1\) if the condition is true and \(0\) otherwise. The condition \( p \in T \) (or \( n \in T \)) denotes the presence of a positive (or negative) keyphrase in the text \( T \).
For sentiment analysis, we predict the sentiment of the text as either positive or negative and anticipate a rise if the text is positive and a fall if it is negative, in alignment with baseline settings. However, since the EDT dataset contains only textual data, we do not employ time-based methods on it. Additionally, for our factor-based methods, we execute the entire SKGP process on benchmark datasets but exclude the use of the $TimeTemplate$ when assessing the EDT dataset. For a fair comparison, we ensure consistency in the processing of baselines according to their original settings. We employ gpt-3.5-turbo-1106, gpt-4, and gpt-4-1106-preview through the API\footnote{\url{https://platform.openai.com/docs/models}} for factor-based methods. The parameters are set as follows: 1) the window size \(t\) is 5; 2) the number of keyphrases and factors \(k\) is 5; 3) the batch size is 64 for BERT series models and 5 for GPT series models. Our experiments are conducted with an NVIDIA RTX A6000 GPU. The details of the pre-trained models used for the baselines are summarized in Appendix \ref{appendix:model_details}.
\begin{table*}
\centering

\caption{Results for LLMFactor and other baselines. The ACC. is shown in percentages (\%). Bold text indicates the best results, while underlined text indicates the second-best. We omit the comparison of time-based methods for the EDT dataset due to its lack of time-series data.}\label{tbl:total_result}
\begin{adjustbox}{width=\textwidth, center}
\begin{tabular}{cc|cc|cc|cc|cc} 
\hline
\multirow{2}{*}{\textbf{Method}}& \multirow{2}{*}{\textbf{Model}}& \multicolumn{2}{c|}{\textbf{StockNet}} & \multicolumn{2}{c|}{\textbf{CMIN-US}}    & \multicolumn{2}{c|}{\textbf{CMIN-CN}} & \multicolumn{2}{c}{\textbf{EDT}}                                      \\
                                 &                                                                        & \textit{ACC.}       & \textit{MCC.}         & \textit{ACC.}           & \textit{MCC.}
                                 & \textit{ACC.}       & \textit{MCC.}         & \textit{ACC.}           & \textit{MCC.}
                                 \\
\cline{1-10}
\multirow{7}{*}{Keyphrase-based}         & Promptrank \textit{ (ACL2023)}                                                            & 51.24          & 0.010                      & 53.28                      & 0.001 
& 50.21          & -0.003                      & 51.78                      & 0.014 
\\
                                 & KeyBERT                                                             & 51.95                  & 0.012                      & 53.40                      & 0.009
                                 & 50.23                  & -0.004                      & 51.84                      & 0.002 
\\
                                 & YAKE                                                             & 51.91          & 0.005                      & 53.13                      & 0.001 
& 50.20          & 0.001                     & 51.88                      & 0.004
\\
                                 & TextRank                                                             & 51.00                  & 0.003                      & 53.99                     & 0.060
                                 & 50.38                  & 0.006                      & 51.76                      & 0.003
\\
                                 & TopicRank                                                             & 51.92          & 0.008                      & 53.75                      & 0.034 
& 50.26          & -0.002                      & 51.80                      & 0.000
\\
                                 & SingleRank                                                             & 50.32                  & 0.005                     & 53.33                      & 0.004
                                 & 50.29                  & 0.002                      & 51.85                      & 0.004
\\
                                 & TFIDF                                                             & 51.86          & 0.001                      & 53.71                      & 0.018 
& 50.27          & -0.002                      & 51.86                      & 0.017 
\\
\cline{1-10}
\multirow{7}{*}{Sentiment-based} & \begin{tabular}[c]{@{}l@{}}EDT\textit{ (ACL2021)} \end{tabular}    & 40.31                     & -0.066                   & 49.86                   & -0.004 & 40.00                     & 0.021                   & \textbf{75.67}                   & 0.026                     
\\
                                 & \begin{tabular}[c]{@{}l@{}}FinGPT\end{tabular}    & 54.91                     & 0.083                   & 59.98           & 0.182
                                 & 55.78                     & 0.120                   & 53.86           & 0.035 
\\
                                 & \begin{tabular}[c]{@{}l@{}}GPT-4-turbo\end{tabular}    & 53.56                     & 0.060                   & 64.61           & 0.284
                                 & 56.94                     & 0.135                   & 55.37           & 0.057 
\\
                                 & \begin{tabular}[c]{@{}l@{}}GPT-4\end{tabular}    & 53.88                     & 0.062                   & 62.18           & 0.260
                                 & 56.96                     & 0.136                  & 50.94           & 0.031 
\\
                                 & \begin{tabular}[c]{@{}l@{}}GPT-3.5-turbo\end{tabular}    & 52.31                     & 0.044                   & 56.10           & 0.156
                                 & 56.68                     & 0.124                   & 54.34           & 0.040 
\\
                                 & \begin{tabular}[c]{@{}l@{}}RoBERTa\end{tabular}    & 54.46                     & 0.088                   & 57.75           & 0.138
                                 & 52.24                     & 0.064                   & 53.66           & 0.029 
\\
                                 & \begin{tabular}[c]{@{}l@{}}FinBERT\end{tabular}    & 55.42                     & 0.111                   & 58.26           & 0.158
                                 & 55.98                     & 0.121                   & 54.98          & 0.043 
\\
\cline{1-10}
\multirow{2}{*}{Time-based} & \begin{tabular}[c]{@{}l@{}}CMIN \textit{(ACL 2023)}\end{tabular}    & 62.69                     & 0.209                   & 53.43                   & 0.046           
 & 55.28                     & 0.111                   & -                   & - 
 \\
                                 & \begin{tabular}[c]{@{}l@{}}StockNet \textit{( ACL 2018)}\end{tabular}    & 58.23                     & 0.081                   & 52.46           & 0.022
                                 & 54.53                     & 0.045                   & -           & -\\
\cline{1-10}

\multirow{3}{*}{Factor-based(ours)} 
& \begin{tabular}[c]{@{}l@{}}
$\textbf{LLMFactor}_{GPT-4-turbo}$
\end{tabular}    & \uline{65.81}                     & \uline{0.228}                   & 61.71                   & 0.228           
 & \textbf{60.59}                     & \textbf{0.245}                   & 59.09                   & \uline{0.082}
 \\
& \begin{tabular}[c]{@{}l@{}}
$\textbf{LLMFactor}_{GPT-4}$
\end{tabular}    & \textbf{66.32}                     & \textbf{0.238}                   & \uline{65.26}                   &\uline{0.284}           
 & \uline{57.16}                     & \uline{0.196}                   & \uline{60.83}                   & \textbf{0.105} 
 \\
 & \begin{tabular}[c]{@{}l@{}}
$\textbf{LLMFactor}_{GPT-3.5-turbo}$
\end{tabular}    & 57.59                     & 0.145                   & \textbf{66.42}                   & \textbf{0.288}           
 & 56.11                     & 0.139                   & 58.11                   & 0.097 
\\      
\hline
\end{tabular}
\end{adjustbox}
\end{table*}

\begin{table*}
\centering
\caption{The ablation study results for LLMFactor, summarizing average values across $LLMFactor_{GPT-4-turbo}$, $LLMFactor_{GPT-4}$, and $LLMFactor_{GPT-3.5-turbo}$. The ACC. is expressed in percentages (\%), with the highest performance bolded for emphasis. The analysis excludes price data comparisons in the EDT dataset due to the absence of time-series data.}\label{tbl:ablation_layer}
\small
\begin{adjustbox}{width=\textwidth, center}

\begin{tabular}{cc|cc|cc|cc|cc} 
\hline
\multirow{2}{*}{\textbf{Method}}& \multirow{2}{*}{\textbf{Model}}& \multicolumn{2}{c|}{\textbf{StockNet}} 
& \multicolumn{2}{c|}{\textbf{CMIN-US}} 
& \multicolumn{2}{c|}{\textbf{CMIN-CN}} 
& \multicolumn{2}{c}{\textbf{EDT}} \\
                                 &                                                                        & \textit{ACC.}       & \textit{MCC.} 
                                 & \textit{ACC.}       & \textit{MCC.} 
                                 & \textit{ACC.}       & \textit{MCC.} 
                                 & \textit{ACC.}       & \textit{MCC.} \\
\cline{1-10}
\multirow{3}{*}{$LLMFactor$    }  &  \textit{Price}   & 52.16                     &  0.041
& 55.59                     &  0.135
& 51.76                     &  0.048
& -                     &  -
\\
   &  +\textit{Factor}   & 58.04                     &  0.166
   & 61.68                     &  0.241
   & 55.71                     &  0.119
   & 55.93                     &  0.077\\
    & $\textbf{+\textit{Factor}+\textit{Relation}}$    & \textbf{63.24}                     & \textbf{0.203}   
    & \textbf{64.46}                     &  \textbf{0.267}
    & \textbf{57.96}                     &  \textbf{0.194}
    & \textbf{59.35}                     &  \textbf{0.095}\\
\hline
\end{tabular}
\end{adjustbox}
\end{table*}
\begin{figure*}
  \centering
  \includegraphics[width=\linewidth]{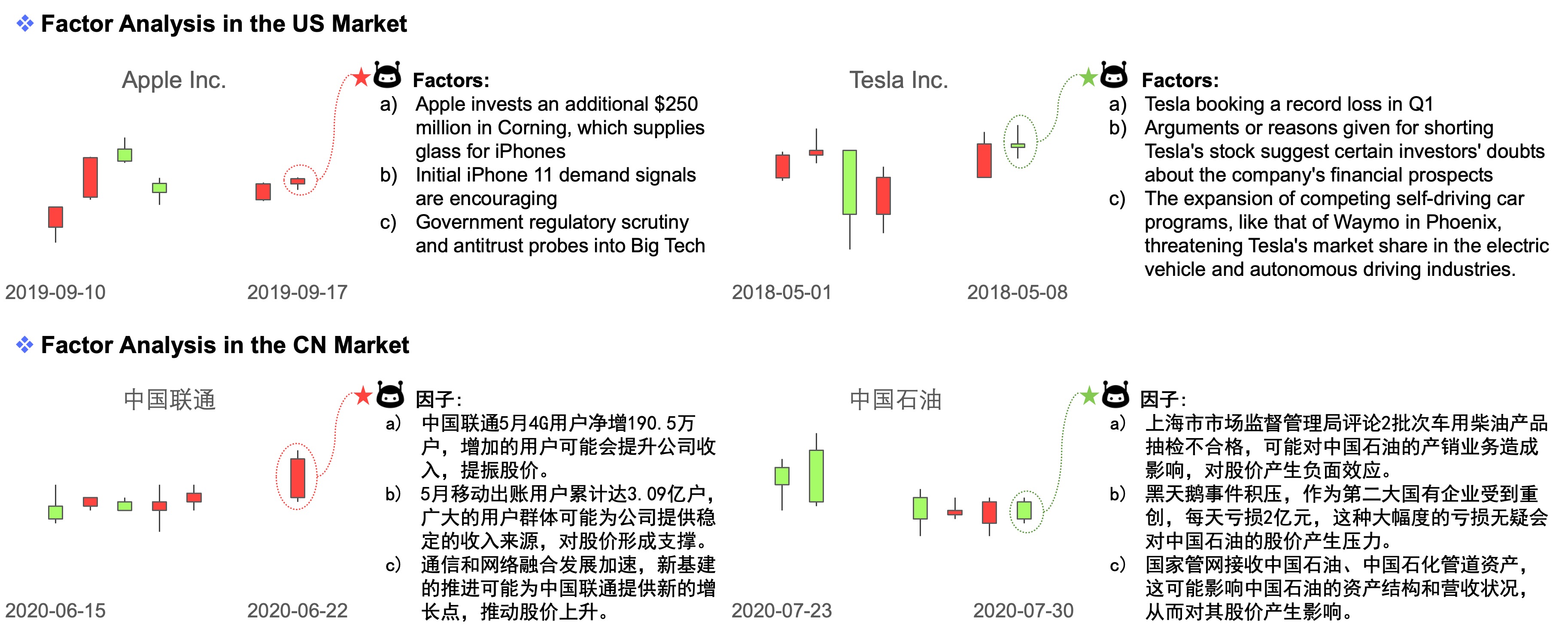}
  \caption{Analyzing the impact of LLMFactor on US and CN markets: LLMFactor is capable of identifying key factors that influence daily fluctuations in stock prices for individual stocks.}
  \label{fig: casestudy}
\end{figure*}
\subsection{Results}
Table \ref{tbl:total_result} presents the results for LLMFactor and various baselines across four datasets. Among all evaluated approaches, our factor-based methods surpass the others, with time-based, sentiment-based, and keyphrase-based methods following in performance. Notably, LLMFactor achieves superior performance over the SOTAs, with improvements of \textbf{2.9\%}, \textbf{0.4\%}, \textbf{11\%}, and \textbf{4.8\%} in MCC across the four datasets, respectively.

The minimal performance differences among keyphrase-based models suggest their limited effectiveness in forecasting stock movements using keyphrases, as keyphrases primarily summarize the main content of the text rather than provide actionable insights on stock prices. Additionally, sentiment-based methods demonstrate varying performance across different models. LLMs like GPT-4 and GPT-4-turbo excel in identifying sentiment within the text. Although the EDT model achieves the highest accuracy on its specific dataset, its low MCC indicates imbalanced model performance. Unlike sentiment-based methods geared towards efficiently extracting relevant sentiment, time-based methods focus on integrating multiple data types. Since time-based methods consider the text in its entirety without detailed analysis, their performance is on par with that of sentiment-based methods, which may lack temporal context but include detailed textual information. However, our LLMFactor model stands out among other SOTAs due to the SKGP technique, which not only identifies significant factors influencing stock prices from textual data but also incorporates relation and temporal information, thereby enhancing its ability to filter out irrelevant content and provide a more comprehensive analysis.

Analyzing average scores from diverse datasets with the use of three factor-based models, our LLMFactor demonstrates an average accuracy of over 63\% and an average MCC of over 0.2 for both the StockNet and CMIN-US datasets, tailored to the US market. However, its performance slightly declines on the CMIN-CN dataset, focused on the CN market, with an average accuracy of 58\% and an MCC of 0.19. This variation indicates that the GPT series models are potentially more proficient in processing English-language text. The EDT dataset, which consists solely of stock-related news without historical price data, poses additional challenges, resulting in a marginally lower average accuracy of 59\% and an MCC of 0.1 for the LLMFactor. The absence of historical price information in the EDT dataset likely diminishes the LLMFactor's effectiveness, underscoring the importance of comprehensive data for financial market analysis.

\subsection{Data Analysis}
\subsubsection{Ablation Study}
We conduct an ablation study on the LLMFactor to demonstrate the improvement contributed by each layer. The first layer utilizes only the $TimeTemplate$, which incorporates past price movements. The second layer incorporates additional $factors$ extracted, and the third layer further includes $relations$ obtained. We average the scores for ACC and MCC from $LLM_{GPT-4-turbo}$, $LLM_{GPT-4}$, and $LLM_{GPT-3.5-turbo}$, as shown in Table \ref{tbl:ablation_layer}. The price layer accounts for approximately 86\% and 32\% of the total performance in ACC and MCC, respectively, while the factor layer contributes an improvement of 9\% in ACC and 46\% in MCC. The relation layer leads to a 5\% and 22\% enhancement in ACC and MCC, respectively. These results imply that the factor layer contributes most significantly to the overall performance of the proposed LLMFactor.

We also conduct experiments on different types of $FactorTemplate$, as shown in Appendix \ref{appendix:ablation_type}.
\subsubsection{Case Study}
In this section, we highlight the practical effectiveness of LLMFactor through an illustrative application. Figure \ref{fig: casestudy} presents a case study where we apply factor analysis to selected stocks in the US and CN markets. For instance, on September 17, 2019, Apple Inc.'s stock (AAPL) experienced an increase, attributed to significant developments such as a \$250 million investment in its supplier, Corning Incorporated, and the positive initial demand for the iPhone 11. Conversely, on May 8, 2018, Tesla Inc.'s stock (TSLA) experienced a decline, mainly due to a record quarterly loss, growing negative concerns among investors, and the increasing competition in the autonomous and electric vehicle markets. This case study demonstrates LLMFactor's ability to effectively integrate background knowledge of company affiliations with both historical news and price data. By analyzing these diverse data sources, LLMFactor significantly enhances the interpretability of stock market dynamics.

\section{Conclusion}
In this study, we introduce LLMFactor, an innovative framework designed to predict and elucidate stock market trends accurately. At the core of LLMFactor is the Sequential Knowledge-Guided Prompting (SKGP) strategy, which integrates background knowledge, stock-related factors, and temporal data to forecast stock movements. Through rigorous testing on four benchmark datasets, LLMFactor has proven its superiority over SOTAs that rely on keyphrases, sentiment analysis, and multimodal data inputs. The application of factor analysis highlights LLMFactor's novelty and effectiveness, establishing it as a powerful tool for financial analysis. This research represents a significant advancement in leveraging LLMs for transparent and explainable financial forecasting.
\newpage
\section{Limitations}
We recognize three main limitations: 
\begin{itemize}[noitemsep,leftmargin=*,align=left]
    \item Although LLMFactor's foundation relies on extensively discussed factors, converting temporal data into text format remains vital for financial forecasting. Future studies could explore additional methods for this conversion and their integration into LLMs.
    \item Due to the variable nature of LLM responses, replicating our experimental results precisely is challenging. In future research, we aim to explore further methodologies to improve reproducibility.
    \item Evaluating LLMFactor against benchmark datasets enables comparison with leading models, but it's essential to carefully assess the quality of factors extracted from texts of various lengths and types. This research marks the beginning of a novel approach to applying factors in finance. Moving forward, we will concentrate on enhancing the quality of these factors.
\end{itemize}
\section*{Acknowledgements}
We sincerely thank the anonymous reviewers for their reviews. This work was supported by JST SPRING, Grant Number JPMJSP2108, and JST-PRESTO, Grant Number JPMJPR2267.

\bibliography{anthology,custom}
\appendix
\section{Templates of SKGP}
\label{appendix:templates of skgp}
The fundamental templates employed at each step of SKGP are detailed in Table \ref{tbl:appendix_template_en} for English and Table \ref{tbl:appendix_template_cn} for Chinese. In the final step of SKGP, we compile the responses from previous steps to formulate prompts for LLMs. Nonetheless, due to the variable nature of LLM responses, the exact reproducibility of results cannot be guaranteed. Instead, we demonstrate the overall superior performance of the LLMFactor.

\section{Details of Pre-trained Models}
\label{appendix:model_details}
We employ the pke \cite{boudin-2016-pke} and pke\_zh \cite{pke_zh} toolkits to access keyphrase-based methodologies, ensuring alignment with established baselines. Additionally, configuration details for sentiment-based models are outlined in Table \ref{tbl:model details}, provided by HuggingFace. The models specified at the top are designated for English texts, whereas the ones at the bottom cater to Chinese texts.

\begin{table}[H]
\centering
\caption{Details of pre-trained models}\label{tbl:model details}
\begin{adjustbox}{width=\linewidth}
\begin{tabular}{ccc}
\hline

\textbf{Model}                            & \textbf{HuggingFace Key}                              & \textbf{Model Size}  \\ \hline \hline
\multicolumn{1}{c|}{\multirow{2}{*}{FinBERT}} 
& \multicolumn{1}{c|}{ProsusAI/finbert} 
& 
438MB
\\
\multicolumn{1}{c|}{}
&
\multicolumn{1}{c|}{bardsai/finance-sentiment-zh-base}
&
409MB
\\
\cline{1-3}
\multicolumn{1}{c|}{\multirow{2}{*}{RoBERTa}} 
& \multicolumn{1}{c|}{soleimanian/financial-roberta-large-sentiment} 
& 
1.42GB
\\
\multicolumn{1}{c|}{}
&
\multicolumn{1}{c|}{IDEA-CCNL/Erlangshen-Roberta-110M-Sentiment}
&
409MB
\\
\cline{1-3}
\multicolumn{1}{c|}{\multirow{2}{*}{FinGPT}} 
& \multicolumn{1}{c|}{FinGPT/fingpt-sentiment\_llama2-13b\_lora} 
& 
14.36GB
\\
\multicolumn{1}{c|}{}
&
\multicolumn{1}{c|}{oliverwang15/FinGPT\_ChatGLM2\_Sentiment\_Instruction\_LoRA\_FT}
&
7.88GB
\\
\cline{1-3}

\end{tabular}

\end{adjustbox}
\end{table}

\section{Comparative Experimental Results of Factor Templates}
\label{appendix:ablation_type}
The experimental results for various factor templates, utilizing three factor-based models: $LLMFactor_{GPT-4-turbo}$, $LLMFactor_{GPT-4}$, and $LLMFactor_{GPT-3.5-turbo}$, are showcased in Table \ref{tbl:ablation_type_en} for English templates and in Table \ref{tbl:ablation_type_cn} for Chinese templates. The datasets tested include CMIN-US and CMIN-CN. Within the English templates, the last template achieves the highest performance, while among the Chinese templates, the initial template emerges as the most effective. The consistent effectiveness of LLMFactor across different templates underscores its robustness in predicting stock market movements.
\begin{table*}
\centering
\caption{SKGP Templates in English}\label{tbl:appendix_template_en}
\Huge
\renewcommand{\arraystretch}{1.3}
\begin{adjustbox}{width=15cm, center}
\begin{tabular}{c|c} 
\hline
\multirow{1}{*}{\textbf{Step}}&
\multirow{1}{*}{\textbf{EN-Template}}
\\
\cline{1-2}
Step1
&
Please fill in the blank and return a complete sentence: $stock_{target}$ and $stock_{match}$ are most likely in a \_\_\_ relationship.
\\
\cline{1-2}
Step2
&
Please extract the top $k$ factors that may affect the stock price of $stock_{target}$ from the following news.
\\
\cline{1-2}
\multirow{9}{*}{Step3}
&
Based on the following information, please judge the direction of the stock price from rise/fall, fill in the blank and give reasons. \\

&
These are the main factors that may affect this stock's price recently: $factor$.
\\

&
These are the connections between the companies that have appeared in the news: $relation$.
\\

&
On $date_{i-5}$, the stock price of $stock_{target}$ $f{(\hat{P_{i-5})}}$.
\\

&
On $date_{i-4}$, the stock price of $stock_{target}$ $f{(\hat{P_{i-4})}}$.
\\

&
On $date_{i-3}$, the stock price of $stock_{target}$ $f{(\hat{P_{i-3})}}$.
\\

&
On $date_{i-2}$, the stock price of $stock_{target}$ $f{(\hat{P_{i-2})}}$.
\\

&
On $date_{i-1}$, the stock price of $stock_{target}$ $f{(\hat{P_{i-1})}}$.
\\

&
On $date_i$, the stock price of $stock_{target}$ will \_\_\_.
\\
\hline
\end{tabular}
\end{adjustbox}
\end{table*}
\begin{table*}
\centering
\caption{SKGP Templates in Chinese}\label{tbl:appendix_template_cn}
\renewcommand{\arraystretch}{1.1}
\begin{adjustbox}{width=15cm, center}
\tiny
\begin{tabular}{c|c} 
\hline
\multirow{1}{*}{\textbf{Step}}&
\multirow{1}{*}{\textbf{CN-Template}}
\\
\cline{1-2}
Step1
&
\begin{CJK*}{UTF8}{gbsn}
请填空并返回完整的句子: \textit{$\text{stock}_{\text{target}}$}
\end{CJK*}
\begin{CJK*}{UTF8}{gbsn}
和
\textit{$\text{stock}_{\text{match}}$}
\end{CJK*}
\begin{CJK*}{UTF8}{gbsn}最可能是\_\_\_关系。    
\end{CJK*}
\\
\cline{1-2}
Step2
&
\begin{CJK*}{UTF8}{gbsn}
请从以下新闻中提取可能影响\textit{$\text{stock}_{\text{target}}$}
\end{CJK*}
\begin{CJK*}{UTF8}{gbsn}
股价的前
\textit{{\text{k}}}
个因素。
\end{CJK*}
\\
\cline{1-2}
\multirow{9}{*}{Step3}
&
\begin{CJK*}{UTF8}{gbsn}
根据以下信息，请判断股票价格是上涨还是下跌，填写在空白处并给出理由。
\end{CJK*}
\\

&
\begin{CJK*}{UTF8}{gbsn}
这些是最近可能影响该股票价格的主要因素: \textit{ \text{factor}}
\end{CJK*}
\\

&
\begin{CJK*}{UTF8}{gbsn}
这些是新闻中出现过的公司之间的关系: \textit{ \text{relation}}
\end{CJK*}
\\

&
\begin{CJK*}{UTF8}{gbsn}
在\textit{$\text{date}_{\text{i-5}}\text{,}$}
\end{CJK*}
\begin{CJK*}{UTF8}{gbsn}
\textit{$\text{stock}_{\text{target}}$}
\end{CJK*}
\begin{CJK*}{UTF8}{gbsn}
的股价\textit{ \text{f}(\text{$\hat{\text{P}_{\text{i-5}}}$})\text{。}}
\end{CJK*}
\\

&
\begin{CJK*}{UTF8}{gbsn}
在\textit{$\text{date}_{\text{i-4}}\text{,}$}
\end{CJK*}
\begin{CJK*}{UTF8}{gbsn}
\textit{$\text{stock}_{\text{target}}$}
\end{CJK*}
\begin{CJK*}{UTF8}{gbsn}
的股价\textit{ \text{f}(\text{$\hat{\text{P}_{\text{i-4}}}$})\text{。}}
\end{CJK*}
\\

&
\begin{CJK*}{UTF8}{gbsn}
在\textit{$\text{date}_{\text{i-3}}\text{,}$}
\end{CJK*}
\begin{CJK*}{UTF8}{gbsn}
\textit{$\text{stock}_{\text{target}}$}
\end{CJK*}
\begin{CJK*}{UTF8}{gbsn}
的股价\textit{ \text{f}(\text{$\hat{\text{P}_{\text{i-3}}}$})\text{。}}
\end{CJK*}
\\

&
\begin{CJK*}{UTF8}{gbsn}
在\textit{$\text{date}_{\text{i-2}}\text{,}$}
\end{CJK*}
\begin{CJK*}{UTF8}{gbsn}
\textit{$\text{stock}_{\text{target}}$}
\end{CJK*}
\begin{CJK*}{UTF8}{gbsn}
的股价\textit{ \text{f}(\text{$\hat{\text{P}_{\text{i-2}}}$})\text{。}}
\end{CJK*}
\\

&
\begin{CJK*}{UTF8}{gbsn}
在\textit{$\text{date}_{\text{i-1}}\text{,}$}
\end{CJK*}
\begin{CJK*}{UTF8}{gbsn}
\textit{$\text{stock}_{\text{target}}$}
\end{CJK*}
\begin{CJK*}{UTF8}{gbsn}
的股价\textit{ \text{f}(\text{$\hat{\text{P}_{\text{i-1}}}$})\text{。}}
\end{CJK*}
\\

&
\begin{CJK*}{UTF8}{gbsn}
在\textit{$\text{date}_{\text{i}}\text{,}$}
\end{CJK*}
\begin{CJK*}{UTF8}{gbsn}
\textit{$\text{stock}_{\text{target}}$}
\end{CJK*}
\begin{CJK*}{UTF8}{gbsn}
的股价将\_\_\_。
\end{CJK*}

\\
\hline
\end{tabular}
\end{adjustbox}
\end{table*}
\begin{table*}
\centering
\caption{Experimental Results for Different Factor Templates in English}
\label{tbl:ablation_type_en}
\Huge
\renewcommand{\arraystretch}{1.3}
\begin{adjustbox}{width=15cm, center}
\begin{tabular}{l|cc|cc|cc} 
\hline
\multirow{2}{*}{\textbf{Factor Template}}& \multicolumn{2}{c|}{\textbf{LLMFactor$_{GPT-4-turbo}$}}
& \multicolumn{2}{c|}{\textbf{LLMFactor$_{GPT-4}$}}
& \multicolumn{2}{c}{\textbf{LLMFactor$_{GPT-3.5-turbo}$}}\\
 & \textit{ACC.}       & \textit{MCC.}       
 & \textit{ACC.}       & \textit{MCC.}
 & \textit{ACC.}       & \textit{MCC.}
 \\
\cline{1-7}
Please extract the top $k$ factors that may affect the stock price of $stock_{target}$ from the following news& 61.71                     &  0.228      & 65.26                     &  0.284 & 66.42                     &  0.288            \\
Please identify the primary top $k$ factors influencing $stock_{target}$'s stock price based on the news provided& 66.98                     &  0.292     & 64.83                     & 0.246 & 65.26                     &  0.298             \\
Please analyze the provided news and pinpoint the top $k$ major factors impacting the stock price of $stock_{target}$   & 65.29                     & 0.295    & 69.21                     & 0.312 & 66.56                     & 0.293               \\
\hline
\end{tabular}
\end{adjustbox}
\end{table*}
\begin{table*}
\centering
\caption{Experimental Results for Different Factor Templates in Chinese}
\label{tbl:ablation_type_cn}
\small
\renewcommand{\arraystretch}{1.2}
\begin{adjustbox}{width=15cm, center}
\begin{tabular}{l|cc|cc|cc} 
\hline
\multirow{2}{*}{\textbf{Factor Template}}& \multicolumn{2}{c|}{\textbf{LLMFactor$_{GPT-4-turbo}$}}
& \multicolumn{2}{c|}{\textbf{LLMFactor$_{GPT-4}$}}
& \multicolumn{2}{c}{\textbf{LLMFactor$_{GPT-3.5-turbo}$}}\\
 & \textit{ACC.}       & \textit{MCC.}       
 & \textit{ACC.}       & \textit{MCC.}
 & \textit{ACC.}       & \textit{MCC.}
 \\
\cline{1-7}
\begin{CJK*}{UTF8}{gbsn}
请从以下新闻中提取可能影响$stock_{target}$股价的前$k$个因素
\end{CJK*}& 60.59                     &  0.245      & 57.16                     &  0.196 & 56.11                     &  0.139           \\
\begin{CJK*}{UTF8}{gbsn}
根据提供的新闻，请识别出影响$stock_{target}$股价的主要$k$个因素
\end{CJK*}& 57.92                     &  0.147     & 64.79                     & 0.109 & 65.13                     &  0.139             \\
\begin{CJK*}{UTF8}{gbsn}
请分析所提供的新闻并找出影响$stock_{target}$股价的前$k$个主要因素
\end{CJK*}   & 64.29                     & 0.160    & 59.33                     & 0.053 & 59.49                     & 0.033             \\
\hline
\end{tabular}
\end{adjustbox}
\end{table*}

\end{document}